\PassOptionsToPackage{table}{xcolor}
\documentclass[acmsmall,nonacm]{acmart}
\AtBeginDocument{%
  }
\usepackage[normalem]{ulem}
\usepackage{graphicx}
\usepackage{subcaption}

\usepackage{wrapfig}
\usepackage{multirow}
\usepackage{array}         
\usepackage{ragged2e}      
\usepackage{makecell} 
\usepackage{longtable}
\usepackage{array}
\usepackage{multirow}

\usepackage{makecell}

\usepackage{tcolorbox}
\usepackage{tikz}
\usepackage{adjustbox}
\usepackage{listings}
\newcolumntype{L}[1]{>{\raggedright\let\newline\\\arraybackslash}p{#1}}  

\setcopyright{none}

\copyrightyear{2025}

\begin{document}

\title[Cognitive Guard Rails for Open-World sUAS Decisions]{Cognitive Guardrails for Open-World Decision Making in Autonomous Drone Swarms}

\author{Jane Cleland-Huang}
\email{JaneHuang@nd.edu}
\orcid{0000-0001-9436-5606}
\affiliation{%
  \institution{Computer Science and Engineering, University of Notre Dame}
  \city{Notre Dame}
  \state{Indiana}
  \country{USA}
}

\author{Pedro Antonio Alarcon Granadeno}
\email{[palarcon@nd.edu]}
\orcid{0009-0006-7829-7088}
\affiliation{%
  \institution{Computer Science and Engineering, University of Notre Dame}
  \city{Notre Dame}
  \state{Indiana}
  \country{USA}
}

\author{Arturo Miguel Russell Bernal}
\email{[arussel8@nd.edu]}
\orcid{[0009-0009-2902-5766]}
\affiliation{%
  \institution{Computer Science and Engineering, University of Notre Dame}
  \city{Notre Dame}
  \state{Indiana}
  \country{USA}
}
\author{Demetrius Hernandez}
\email{[dhernan7@nd.edu]}
\orcid{[to be filled later]}
\affiliation{%
  \institution{Computer Science and Engineering, University of Notre Dame}
  \city{Notre Dame}
  \state{Indiana}
  \country{USA}
}

\author{Michael Murphy}
\email{[email@nd.edu]}
\orcid{[to be filled later]}
\affiliation{%
  \institution{Computer Science and Engineering, University of Notre Dame}
  \city{Notre Dame}
  \state{Indiana}
  \country{USA}
}

\author{Maureen Petterson}
\email{[mpetters@nd.edu]}
\orcid{[0009-0008-8801-0751]}
\affiliation{%
  \institution{Computer Science and Engineering, University of Notre Dame}
  \city{Notre Dame}
  \state{Indiana}
  \country{USA}
}

\author{Walter Scheirer}
\email{[walter.scheirer@nd.edu]}
\orcid{0000-0001-9649-8074}
\affiliation{%
  \institution{Computer Science and Engineering, University of Notre Dame}
  \city{Notre Dame}
  \state{Indiana}
  \country{USA}
}
\renewcommand{\shortauthors}{Cleland-Huang, Granadeno, Russell, Hernandez, Murphy, Petterson, Scheirer}
\thanks{© 2025 by the authors. This work is licensed for arXiv preprint submission.}
\begin{abstract}
Small Uncrewed Aerial Systems (sUAS) are increasingly deployed as autonomous swarms in search-and-rescue and other disaster-response scenarios. In these settings, they use computer vision (CV) to detect objects of interest and autonomously adapt their missions. However, traditional CV systems often struggle to recognize unfamiliar objects in open-world environments or to infer their relevance for mission planning. To address this, we incorporate large language models (LLMs) to reason about detected objects and their implications. While LLMs can offer valuable insights, they are also prone to hallucinations and may produce incorrect, misleading, or unsafe recommendations. To ensure safe and sensible decision-making under uncertainty, high-level decisions must be governed by cognitive guardrails. This article presents the design, simulation, and real-world integration of these guardrails for sUAS swarms in search-and-rescue missions.
\end{abstract}

\ccsdesc[500]{Computer systems organization~Robotic autonomy}
\ccsdesc[500]{Computing methodologies~Robotic planning}
\ccsdesc[500]{Computing methodologies~Planning under uncertainty}
\ccsdesc[500]{Computing methodologies~Scene understanding}
\ccsdesc[500]{Human-centered computing~Collaborative interaction}

\keywords{autonomous swarms, open-world decision making, cognitive guardrails, small uncrewed aerial systems, human-machine teaming}


\maketitle

\section{Motivation}
Swarms of autonomous Small Uncrewed Aerial Systems (sUAS) are increasingly deployed on diverse missions, such as infrastructure inspections, disaster relief, and search-and-rescue (SAR) \cite{next-gen-drones,mishra2020drone,kumar2024uav}. They are well-suited to these operations as they can rapidly cover remote and challenging terrain, perform aerial surveillance, and operate with minimal human intervention. In such scenarios, the sUAS use onboard Computer Vision (CV) and other sensing technologies to detect and classify objects of interest \cite{open-world-cv}. For example, in a SAR scenario, the sUAS might detect objects, such as footprints, clothing, or broken branches that serve as clues in the search. However, CV models tend to be `closed world', meaning that they are trained on specific types of objects, limiting their ability to correctly comprehend a real-world scene with novel objects, activities, and interactions~\cite{sunderhauf2018limits,BoultandScheirer2023}. To help overcome this limitation, we integrate large language models (LLMs) into the sUAS reasoning infrastructure, enabling sUAS' to interpret implications of discovered open-world objects, plan actions, and adapt autonomously in response.

\begin{figure*}
\begin{minipage}{0.5\linewidth}
    \centering
    \includegraphics[width=\linewidth, trim=0cm 1cm 0cm 1cm, clip]{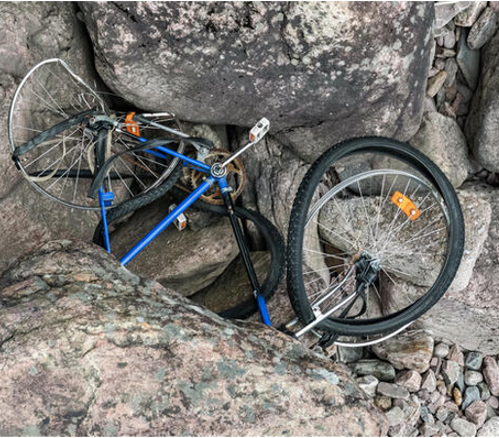}
\end{minipage}
\hfill
\begin{minipage}{0.45\linewidth}
    \scriptsize
    \textbf{Prompt \#1:} “A cyclist is lost in the wilderness. Can you find a person or any signs of a lost person in this picture? Answer in one sentence.”\\[1ex]
    \textbf{Response \#1:} “There is no person visible in the image, but the damaged bicycle lodged between large rocks is a strong sign that a lost or injured cyclist may have been in the area.”\\[3ex]
    \textbf{Prompt \#2:} ``A cyclist \uline{riding a red bike} is lost. Is this bicycle relevant to the search for the cyclist? Answer in one sentence.”\\[1ex]
    \textbf{ Response \#2a: [Erroneous]} ``Yes, this clue is relevant, as the damaged bike suggests a potential accident despite differences from the described red mountain bike.''\\[1ex]
    \textbf{ Response \#2b: [Correct]} ``No, the bike in the image does not match the description of the lost red mountain bike, as it is blue and black with thinner tires and lacks typical mountain bike features.''
\end{minipage}
\caption{LLM-based visual reasoning over an AI-generated scene demonstrates how an sUAS leverages the LLM to guide open-world scene interpretation. However, while it exhibits extensive reasoning capabilities, without clear directives, it tends to draw wrong conclusions (e.g., Response 2a) (Image credit: Valery Yurasov)}
\label{fig:bike}
\end{figure*}

The use of LLMs during search is illustrated in Fig.\ref{fig:bike}, where an sUAS engaging in a Search and Rescue (SAR) mission tentatively identifies a damaged blue bicycle with low confidence and uses an LLM for further analysis \cite{wang2023visionllm}. In response to a general prompt about signs of a lost person (Prompt\#1), the LLM correctly infers that the bike is a relevant clue (Response~\#1). However, when informed that the missing cyclist was riding a red bike (Prompt~\#2), two separate queries produce contradictory responses: one erroneously affirms relevance despite the color mismatch (Response~\#2a), while the other correctly rules it out (Response~\#2b). This illustrates both the potential and the inconsistency of using LLMs to reason about open-world clues. 

This article presents the CAIRN framework, which stands for {\it Context-Aware Inference for Reasoning and plaNning}, and is designed to guide autonomous mission-level decision-making in dynamic and uncertain environments. Like a cairn, a stack of stones that marks a path through uncertain terrain, the system incrementally refines decisions through context-sensitive reasoning. CAIRN combines Bayesian inference with natural language analysis of discovered clues to inform planning and adapt strategies as new information emerges. These decisions are constrained by cognitive guardrails that are designed to ensure all actions remain safe, justified, and aligned with mission goals.
To illustrate the framework, we describe CAIRN within the context of SAR missions, where timely autonomous decisions can be life-saving and uncertainty is pervasive.


\section{Bayesian Models for Contextual Reasoning and Planning} \label{sec:bayesian}
\begin{figure}[t]
    \centering

    \begin{subfigure}[t]{0.48\linewidth}
        \centering
        \includegraphics[width=\linewidth, trim=0cm 0cm 0cm .4cm, clip]{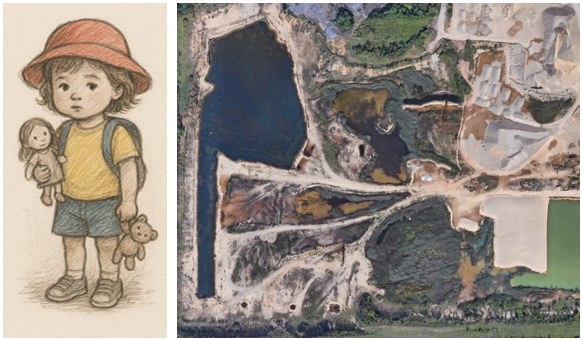}
        \subcaption{A small girl wearing a red hat, yellow shirt, blue shorts, gym shoes, and a backpack, carrying a doll and teddy bear, is lost in a quarry with ponds, sandy trails, and wooded areas. She was last seen northwest of the quarry. Weather is clear and it is daylight.
}\label{fig:lost-girl}
    \end{subfigure}\hfill
    \begin{subfigure}[t]{0.48\linewidth}
        \centering
        \includegraphics[width=\linewidth, trim=0cm 0cm 0cm 0cm, clip]{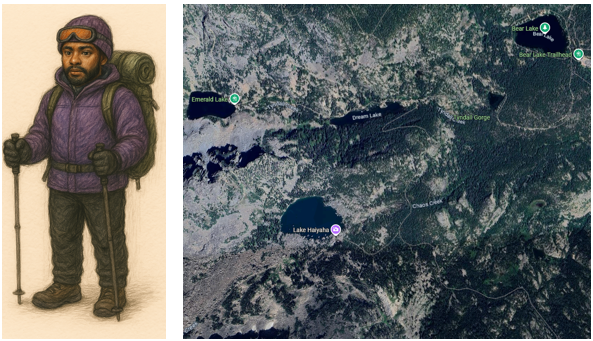}
        \subcaption{An adult male hiker wearing a purple jacket and hat, gray hiking pants, brown boots, and goggles, carrying a backpack and poles, is lost in a mountainous area of the Rockies with steep elevation, lakes, and forests. Weather is clear and it is daylight.
}\label{fig:lost-hiker}
    \end{subfigure}

    \caption{Search-and-rescue strategies are based on Last Known Point (LKP) and time, lost-person descriptions, terrain, weather, and daylight conditions.}
    \label{fig:side-by-side-lost}
\end{figure}

Studies on lost persons have identified common behavioral patterns associated with different profiles. As depicted in Fig. \ref{fig:side-by-side-lost}, a young child will typically remain near their last known location and avoid difficult terrain, whereas a healthy adult is more likely to travel farther, follow trails, or seek out prominent features such as ridge-lines or water sources. Environmental factors, including temperature, precipitation, and daylight, also significantly influence movement and behavior. This behavioral knowledge informs the initial planning phase of an SAR mission by guiding the selection of the best search strategy. As the mission unfolds, real-time observations and discovered clues, such as footprints or discarded belongings, can trigger changes or reprioritize regions of interest. In response, sUAS must adapt dynamically. This requires balancing two complementary paradigms of reactive control, which governs immediate responses to environmental stimuli, and human-like reasoning, which supports deliberative, symbolic decision-making at the mission level \cite{ghallab2004automated}. This article focuses on the latter, emphasizing high-level reasoning that aligns with human thought processes, and is conducive to human-machine teaming (HMT) \cite{6697830,cleland2022extending}.

\subsection{Modeling Search Strategies}
\begin{figure}[t]
    \centering
    \includegraphics[width=\linewidth]{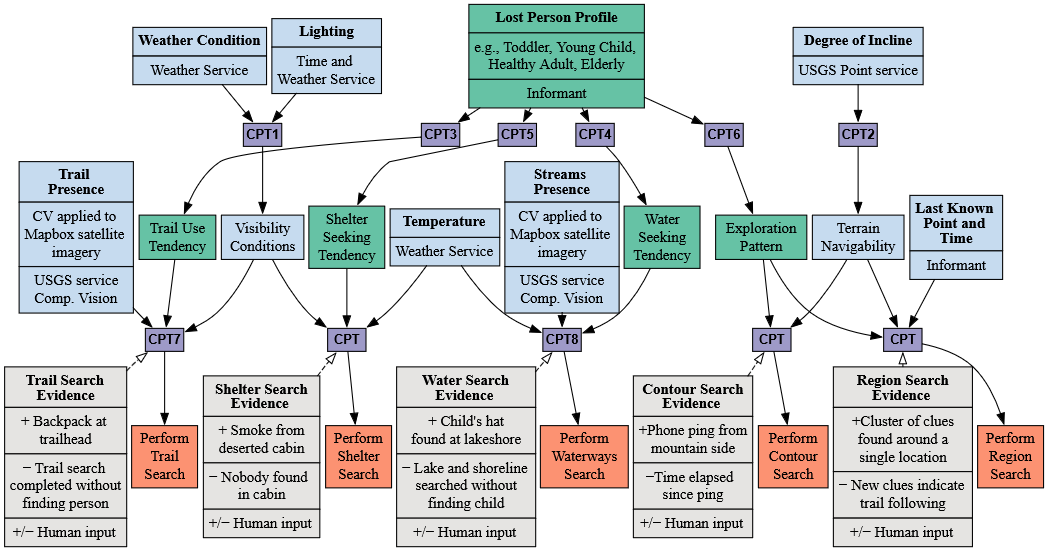}
    \caption{The Bayesian Model computes the probability that each of five search strategies (trail, shelter, waterways, contour, and region) will most likely lead to the missing person. Nodes represent environment and weather (green), hiker profile (blue), and search strategies (orange). Runtime updates to the model are made to the evidence nodes (gray). We populated the Conditional Probability Tables (CPTs) in the model based on documented lost-person behavior patterns, supplemented by principled assumptions informed by SAR domain expertise}
    \label{fig:bayes_model}
\end{figure}

We start by modeling the mission context using a discrete Bayesian model. As shown in Fig.~\ref{fig:bayes_model}, the model includes nodes representing environmental conditions (blue), lost person profiles (green), search strategies (orange), observed evidence (gray), and conditional probability tables (light purple). Given input data on the environment, lost person profile, and emerging evidence, the model infers the most probable high-level search strategy. Although the Bayesian model is primarily intended to guide sUAS decision-making, it can also assist human operators by recommending an effective search strategy.

At the start of each mission environmental data is used to initialize the model. Weather and lighting data are retrieved from real-time meteorological services; terrain characteristics are inferred from GIS data and satellite maps, enhanced by image analysis; while a lost person profile, including a description of the person and the Last Known Point (LKP), is established from informant reports.

This information constitutes the initial evidence \( E_0 \) needed to configure the Bayesian model and to compute a posterior belief over candidate search strategies \( S = \{s_1, s_2, \dots, s_n\} \). Based on this evidence, each strategy \( s_i \) (e.g., Trail Search, Shelter Search, or Waterways Search) is assigned an initial posterior probability \( P_0(s_i \mid E_0) \), which reflects how well that strategy fits the characteristics of the current mission. Specifically:
\begin{equation}\label{eq:1}
P_0(s_i \mid E_0) = \frac{P(E_0 \mid s_i) \cdot P_0(s_i)}{\sum_j P(E_0 \mid s_j) \cdot P_0(s_j)}
\end{equation}
where:
\begin{itemize}
    \item $P_0(s_i)$ is the prior belief in strategy $s_i$, based on the lost person profile and initial mission context;
    \item $P(E_0 \mid s_i)$ is the likelihood of encountering the observed environmental conditions $E_0$ assuming strategy $s_i$ is appropriate (e.g., the presence of trails increases the likelihood of trail-based search);
    \item $P_0(s_i \mid E_0)$ is the updated belief in strategy $s_i$ after observing $E_0$, used to guide early swarm deployment. The distribution is normalized so that $\sum_i P_0(s_i \mid E_0) = 1$.
\end{itemize}

This yields a probability distribution that reflects how well each strategy aligns with current environmental conditions and the known details of the missing person. The dominant strategy is then adopted at the start of the mission.

\subsection{Runtime Belief Updates}
\label{sec:updates}

As the mission progresses, new clues may be detected, tasks completed, environmental conditions changed, and human inputs provided. These serve as evidence to be incorporated into the Bayesian model, reflecting the current state of the search and impacting its strategies as depicted in Table \ref{tab:llm_trace_redhat}. Model updates are performed using the following equation:

\begin{equation} \label{eq:2}
P_t(s_i) = \frac{(1 + \gamma \cdot \delta_{i}) \cdot P_{t-1}(s_i)}{{\sum_j (1 + \gamma \cdot \delta_{j}) \cdot P_{t-1}(s_j)}}
\end{equation}

\noindent
where:
\begin{itemize}
    \item $P_{t-1}(s_i)$ is the prior belief in strategy $s_i$.
    \item $P_t(s_i)$ is the updated belief in strategy $s_i$ after incorporating new evidence at time $t$.
    \item $\gamma$ is the adjustment factor that reflects the strength and direction of the update.
    \item $\delta_{i}$ is an indicator function defined as:
    \[
    \delta_{i} =
    \begin{cases}
        1 & \text{if } i =\ \text{the strategy being directly updated} \\
        0 & \text{otherwise}
    \end{cases}
    \]
\end{itemize}
This equation supports both positive and negative belief adjustments, depending on the evidence type, while ensuring that all updated beliefs remain normalized across the candidate strategies. Below, we describe how this is applied to strengthen and weaken beliefs in a strategy. 

\paragraph{Positive Evidence Updates.}  
Positive evidence primarily arises from discovered clues or sightings of the lost person. The discovery of a potentially relevant clue (e.g., a backpack found near a trailhead) can trigger a shift in strategy (e.g., switching from Region Search to Trail Search), and the supported strategy therefore receives a proportional increase in belief. This adjustment is applied using Eq.~\ref{eq:2}, where $\gamma = \alpha > 0$, and $\alpha$ reflects the strength of the evidence based on both its \textit{relevance} and \textit{confidence}.
We compute $\alpha$ using the following weighted combination:

\begin{equation} \label{eq:3}
\alpha = \left( \lambda \cdot R \right) + \left( (1 - \lambda) \cdot \left( \mu \cdot C_{\text{cv}} + (1 - \mu) \cdot C_{\text{interp}} \right) \right)
\end{equation} where:
\begin{itemize}
    \item $R$ is the relevance score assigned to the clue,
    \item $C_{\text{cv}}$ is the confidence in the clue’s object classification derived from a combination of Open-World Computer Vision and the Visual LLM,
    \item $C_{\text{interp}}$ is the tactical confidence in the clue’s significance,
    \item $\lambda \in [0,1]$ weights the influence of relevance versus confidence,
    \item $\mu \in [0,1]$ balances CV classification and interpretation confidence.
\end{itemize}

Relevance and $C_{\text{interp}}$ values are produced by our LLM pipeline (described in Section \ref{sec:clues}) and mapped to numeric values using a common scale: High $\rightarrow$ 0.8, Medium $\rightarrow$ 0.4, Low $\rightarrow$ 0.1, whereas $C_{\text{cv}}$ is produced by the Computer Vision module, and potentially boosted by the LLM's classification, when CV uncertainty is low.
 While the update targets a specific strategy, all probabilities are adjusted proportionally through normalization to maintain a valid distribution. Strategy updates may also result from environmental changes (e.g., sunset increasing the likelihood of shelter-seeking behavior) or operator input via the user interface, incorporating contextual knowledge not otherwise captured by the model. Details on how relevance and confidence scores are derived from discovered clues are provided in Section~\ref{sec:clues}.

\paragraph{Negative Evidence Updates.}  
Conversely, negative evidence largely arises when a strategy has been followed without results (e.g., most trail segments searched), or when environmental conditions reduce effectiveness of a strategy. For example, we could decrease belief by setting $\gamma = -\beta$, where $\beta \in [0,1)$ reflects the proportion of a strategy’s area that has been covered.  To prevent premature belief decay, this update is deferred until a minimum coverage threshold (e.g., 60\%) has been reached. The belief in a strategy may decrease for other reasons including changes in weather or daylight conditions, or through human input. Again normalization is used to maintain a valid distribution.

\paragraph{Resetting Strategies.}
Finally, as a search proceeds, human operators may decide to expand the search region, or repeat a prior task due to potential movement (e.g., back-tracking) of the lost person. In such cases, strategies can be individually boosted or reduced using Eq. \ref{eq:2}, or can be entirely reset using Eq. \ref{eq:1}.

When updates to the model indicate a shift in search priorities, the Global Planner may redirect sUAS to focus on newly prioritized strategies or tasks, ensuring that search efforts remain aligned with the evolving mission.

\section{Interpreting Clues}
\label{sec:clues}
To operate autonomously, each sUAS must be capable of reasoning about clues detected using onboard computer vision. Once a potential clue is found, the sUAS uses its Local Planner, supported by off-line reasoning services running the LLM pipeline, to decide whether to {\it do nothing}, {\it autonomously adapt its own behavior}, {\it request help from the swarm} for collective action, or defer the decision to a {\it human operator}. Collective decisions are coordinated by the swarm-level Global Planner.

\begin{figure}[ht]
\begin{minipage}[t]{0.402\linewidth}
    \centering
    \includegraphics[width=\linewidth,trim={20 0 30 0},clip]{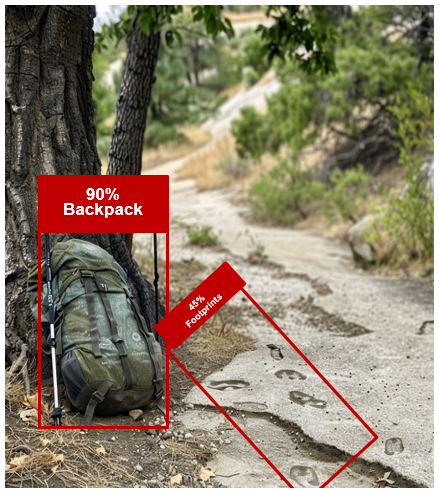}
    \caption{An sUAS discovers a backpack at the trailhead while conducting a trail-based search for a lost hiker. This clue may indicate the person’s entry point, prompting a shift in search strategy or immediate task reassignment.}
    \label{fig:backpack}
\end{minipage}
\hfill
\begin{minipage}[t]{0.55\linewidth}
    \centering
    \includegraphics[width=\linewidth]{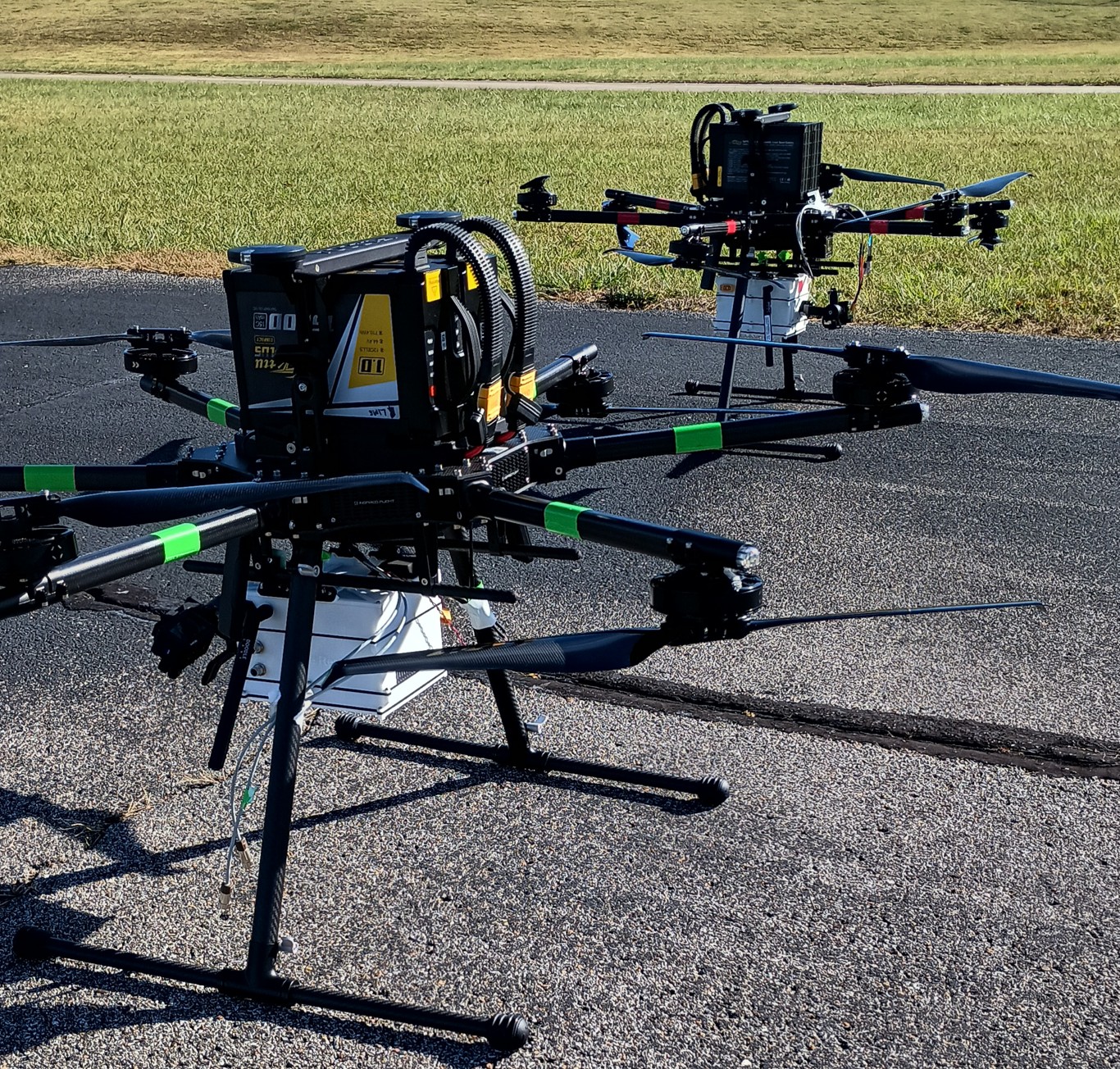}
    \caption{Two hexacopters, each equipped with an Edge Matrix Cube (mCube), await their missions. When assigned a mission, they autonomously configure their internal state machines, take off, and execute their initial task. They evaluate and act upon clues found, and request new tasks when the current task is completed.}
    \label{fig:drones}
\end{minipage}
\end{figure}

\begin{table}[]
    \centering
        \caption{Traditional and Open-World Computer Vision (Stages 1-2), LLM-based Reasoning Stages (3–7), followed by Heuristic steps to engage human, update the Bayesian Model, and take mission actions.  Output values shown in bold and colored (e.g., red or blue) serve as confidence-weighted adjustment factors for clue-based updates (see Section~\ref{sec:updates} for usage). Cognitive Guardrails (depicted as warning signs) are applied either during decision-making or at runtime, in order to ensure mission-alignment.}
    \label{tab:llm_trace_redhat}
    \includegraphics[width=\linewidth]{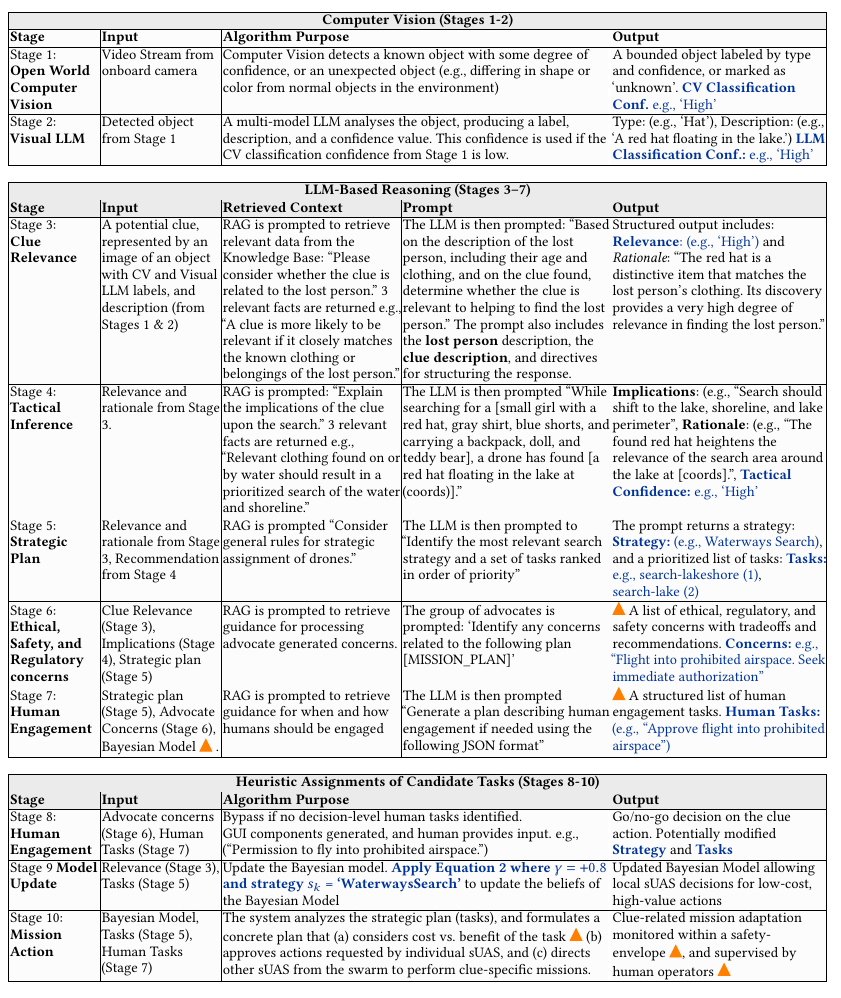}

\end{table}

For example, in Figure~\ref{fig:backpack}, an sUAS conducting a region-based search for the adult hiker (cf., Figure \ref{fig:lost-hiker}) detects a backpack at a trailhead. The sUAS' Local Planner evaluates the clue’s salience and mission context, decides that additional information is needed to determine relevance to the lost person, and initiates a close-up inspection which confirms that the object's appearance matches the description of the hiker's backpack.  This suggests that a trail search would be effective.  However, if the same backpack were found during a search for a missing child, it would carry less significance as it does not match the child's backpack. On the other hand, it could suggest the presence of an unrelated adult, raising additional questions. Addressing such context-dependent clues requires advanced reasoning, as it is infeasible to predefine the implications and appropriate actions for every possible clue-like object.

Therefore, we integrate an LLM into a multi-stage decision pipeline that combines chain-of-reasoning, LangChain orchestration, and retrieval-augmented generation (RAG) \cite{topsakal2023creating}. CAIRN's LLM pipeline is accessible to Local Planners onboard each sUAS and to a system-wide Global Planner. Pipeline execution is triggered when a clue is detected by the onboard Computer Vision system.  This triggers a sequence of LLM prompts, enriched with contextual knowledge retrieved from a curated database that we have compiled using SAR best practices documented in white papers and books (e.g., \cite{SAR-1,SAR-2,SAR-3,SAR-4}).  Our initial knowledge base includes 73 entries, stored as vector embeddings and organized using a hierarchical tag structure to enable fast, targeted retrieval for each stage of the pipeline. For example, two entries providing guidance for tactical planning include: ``If the lost person's belongings are found on a trail, prioritize a directional trail search with emphasis on the downhill path'', or ``If a clue is found in a body of water, rapidly dispatch drones to search the water and the shoreline.''  

The LLM pipeline is depicted in Table \ref{tab:llm_trace_redhat}, illustrated with an example in which a red hat has been found during a search for the lost girl (see Fig. \ref{fig:lost-girl}). 
In Stages 3–5, the system assesses the relevance of the clue (Stage 3), identifies tactical implications (Stage 4), and generates a strategic plan (Stage 5). In Stage 6, ethical, safety, and regulatory concerns associated with the plan are identified and assessed, while Stage 7 determines if the sUAS can act autonomously or if a human operator should be engaged in the decision. If needed, Stage 8 generates and displays the appropriate GUI elements for eliciting human feedback, and waits for a response. Unless the human rejects the clue, the Bayesian model is updated in Stage 9 using Eq.  \ref{eq:2}. Here, $\gamma$ is derived from Eq. \ref{eq:3}, using the relevance score form Stage 3 (`High' = 0.8), the classification confidence from Stage 1 (`High' = 0.8) and the confidence in tactical interpretation from Stage 4 (`High' = 0.8'); with hyperparameters $\lambda$ = 0.5 and $\mu$ = 0.5, the value of $\gamma$ equals 0.8. Note that because the classification confidence from Stage 1 is `High', the classification confidence form Stage 2 gets omitted, for this particular example. The strategy to be directly updated $s_k$ is derived from the output of Stage 5 (i.e., Strategy = `Waterways Search'). Given the LLM's propensity for generating incorrectly formatted outputs, we include automatic detection, repair, and regeneration of outputs as an integral part of our LLM pipeline.  
Finally, in Stage 10, the Local and/or Global Mission Planners, use the task plans generated in Stage 5, and potentially refined by humans in Stage 8, to assign tasks to sUAS or to queue tasks for later execution.

\section{Placing Guardrails around Autonomous Decisions}
\label{sec:guardrails}
While the CAIRN pipeline enables rich, contextual reasoning, it also introduces risks, caused by hallucinations and other LLM-related errors, that could potentially lead to sUAS performing unnecessary or even nonsensical missions. To mitigate these risks, we have embedded several safeguards within the CAIRN pipeline. These include format validation and repair of LLM output, contextual grounding through RAG, cognition-level guardrails that operate during the decision-making process, runtime supervision by human operators, and an operational safety-envelope. This multi-faceted approach is designed to ensure that all enacted decisions are structurally valid, contextually grounded, compliant with cognitive guardrails, and bounded by safety constraints at runtime.

\subsection{Decision-Time Cognitive Guardrails}
Decision-time guardrails are integrated into the decision-making process before any actions are executed. 

\subsubsection{Belief Entropy}
The first cognitive guardrail, applied in Stage 7 of the pipeline, determines when an sUAS (or the sUAS swarm) can act autonomously  on a clue without human supervision. It is based upon the extent to which the planned action aligns with current strategies and beliefs within the Bayesian model \cite{bai2016information,nelly-surprise}.  As illustrated in Table~\ref{tab:entropy-a}, when entropy is low (e.g., Region Search dominates with 65\% probability), the sUAS is allowed to act autonomously within the dominant strategy or may switch into that strategy from a less dominant one. However, switching from a dominant strategy to a less dominant one, or between less-dominant strategies requires human approval. Conversely, when entropy is high, the sUAS may adapt between any strategy that is within a preset threshold of the highest ranked strategy, but must notify the human operator that it has adapted strategies or tasks. In the example depicted in Table \ref{tab:entropy-b}, an sUAS could therefore adapt autonomously across all strategies except for Contour Search, for which human permission would be required. 
Entropy levels in the model reflect both strategic disagreement and uncertainty in interpreting the current clue, as described in Section~\ref{sec:updates}.
\begin{table}[h]
\centering
\caption{The entropy levels of the strategies generated by the Bayesian Model determine when an sUAS can adapt autonomously versus requesting permission from a human operator.}
\label{tab:entropy}

\begin{subtable}[t]{0.45\linewidth}
\centering
\rowcolors{1}{gray!8}{gray!8}
\caption{Entropy is low, indicating strong consensus on Region Search from the Last-Known Point. sUAS may act autonomously under this belief, but switching strategies (e.g., to Trail Search due to a found backpack) requires human confirmation.}
\label{tab:entropy-a}
\begin{tabular}{|L{3.5cm}|c|}
\hline
\textbf{Search Strategy} & \textbf{Probability} \\
\hline
Region Search    & 0.65 \\
Waterways Search & 0.12 \\
Trail Search     & 0.10 \\
Shelter Search   & 0.08 \\
Contour Search   & 0.05 \\
\hline
\end{tabular}
\end{subtable}
\hfill
\begin{subtable}[t]{0.45\linewidth}
\centering
\rowcolors{1}{gray!8}{gray!8}
\caption{Entropy is high, indicating strategic uncertainty with no dominant strategy. sUAS may adapt among strategies within a probability delta below the threshold. They must notify humans to maintain situational awareness.}
\label{tab:entropy-b}
\begin{tabular}{|L{3.5cm}|c|}
\hline
\textbf{Search Strategy} & \textbf{Probability} \\
\hline
Region Search    & 0.23 \\
Waterways Search & 0.22 \\
Trail Search     & 0.22 \\
Shelter Search   & 0.21 \\
Contour Search   & 0.12 \\
\hline
\end{tabular}
\end{subtable}

\end{table}



\subsubsection{Cost-Benefit Analysis}
The second guardrail, applied in Stages 7 and 10, performs a cost-benefit analysis on proposed actions, allowing autonomous execution only when the expected value outweighs the mission cost. For example, an sUAS detects a red object in a tree, but the image is blurry and the object partially occluded. The LLM-pipeline processes the image and concludes that it could be the child's red hat, a potentially relevant clue. It determines that the cost of inspecting the object is low (e.g., approximately a 2 minute detour), and therefore the sUAS proceeds autonomously. If the estimated cost versus benefit of a task exceeds a predefined threshold, the sUAS defers action, and stores the image and geolocation to a task queue for consideration by humans and the Global Planner; either of whom can approve it for reassignment to an available sUAS.

\subsubsection{Ethical and Regulatory Oversight}
The third guardrail, applied in Stage 6, uses runtime advocate personas \cite{hernandez2025advocates} to check for regulatory, ethical, and safety concerns associated with the strategic plan. Each advocate acts as an expert advisor, using an independent LLM-based reasoning agent grounded in existing standards and guidelines. For example, the Safety Controller draws from MIL-STD-882E for hazard assessment \cite{jensen2022selecting}, DO-178C for software assurance \cite{dmitriev2021toward}, and the NASA Human Integration Design Handbook (HIDH) for fail-safe best practices \cite{silva2019implementation}. The Ethical Governor is grounded in GDPR principles as part of a broader ethical framework for harnessing the power of AI in healthcare and beyond \cite{nasir2024ethical}, IEEE P7001 for transparency \cite{ieee2022ieee}, and Red Cross codes for humanitarian conduct \cite{zomignani2020aid}. The Regulatory Auditor relies on FAA Part 107 regulations for sUAS operations \cite{greenwood2020flying}, including constraints on altitude, line-of-sight operation, night flights, and flights over people or moving vehicles, as well as RTCA DO-200B for logging and traceability \cite{marques2017verification}. These advocates evaluate the planned tasks from their own domain perspectives and flag actions that conflict with their governing principles. For example, if the CAIRN planner proposes crossing protected airspace to investigate signs of a capsized boat, the Regulatory Advocate might flag the airspace violation, the Ethical Advocate might see the violation as warranted given a potentially life-and-death situation; and the Safety Advocate might  assess potential route hazards. If consensus cannot be reached, the plan and corresponding concerns, are escalated to the human operator for a decision.

Our prior work \cite{hernandez2025advocates} generated user-facing alerts displayed directly in the GUI; however, with CAIRN, we are developing Advocates as conversational agents, able to debate and summarize issues in a machine-readable format and to determine when concerns need to be elevated for human input.

\begin{figure}
    \centering
    \includegraphics[width=\linewidth]{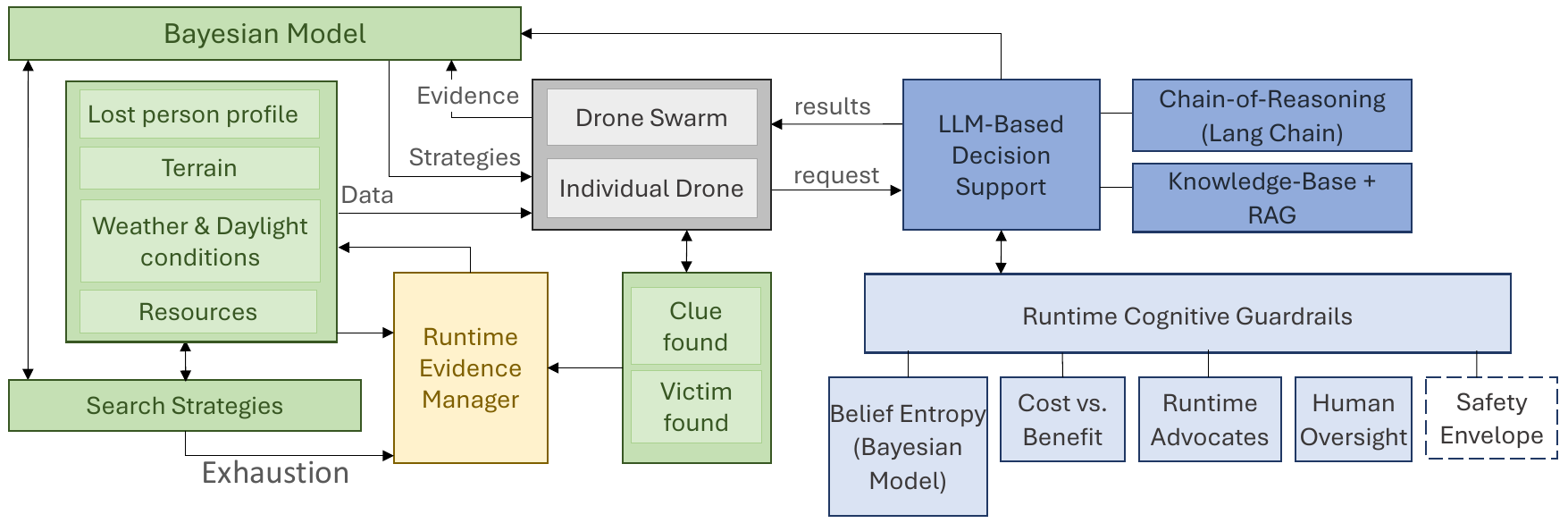}
    \caption{ Decision architecture for autonomous sUAS operations. The Bayesian model provides strategy beliefs based on environmental and mission inputs, while the Runtime Evidence Manager triggers updates as new clues emerge or strategies are exhausted. LLM-based reasoning supports decision-making, guided by five runtime cognitive guardrails. These include belief-based reasoning, cost-benefit analysis, advocate personas, and human oversight, applied during decision time, and a safety envelope, enforced exclusively at runtime to constrain physical execution.}
    \label{fig:decision-architecture}
\end{figure}

\subsection{A Safety Envelope as a Runtime Guardrail}
All actions that are planned and executed at the mission level must operate within a safety envelope that constrains sUAS behavior, independent of mission-level reasoning. The envelope defines hard operational boundaries, such as minimum altitude, maximum range from home position, geofenced exclusions and inclusions, and battery reserve thresholds, that no autonomous behavior may violate, regardless of strategic intent or LLM-generated rationale. Safety envelopes are a standard mechanism in cyber-physical systems (CPS) for monitoring and enforcing non-negotiable constraints and ensuring that system behavior remains within predefined safety limits, even in the presence of adaptive or learning-based components \cite{moore2024safety}. This guardrail therefore ensures that, regardless of reasoning limitations or planning errors, all autonomous behavior remains physically bounded within a safety envelope that enforces strict, non-negotiable constraints.

\subsection{Dual-Phase Human Oversight}
Finally, human-on-the-loop oversight serves as a cross-cutting guardrail that supports safe and accountable autonomy across the mission lifecycle. Unlike the other guardrails, which apply only during decision-making or runtime, this mechanism spans both phases.
During decision-making, human operators are consulted on actions flagged as uncertain, costly, or ethically complex \cite{mape-k-human}. For example, when a detected object is ambiguous or a task carries significant mission risk, the sUAS delays execution and presents relevant information to the human operator. This includes images and locations of clues, action plans and their rationales, and cost estimates for executing the plan. Deferring difficult decisions to humans, balances autonomy with accountability while avoiding unnecessary interruptions for low-risk tasks.

At runtime, operators remain on the loop, maintaining situational awareness of tasks, adaptive decisions, and system intent\cite{endsley2023supporting,next-gen-drones}. To support this, the interface provides visualizations of each sUAS’s assignments, status indicators for autonomous adaptations, and alerts triggered by elevated uncertainty, unresolved trade-offs, or policy conflicts. Operators can inspect decision rationales at any time, enabling proactive intervention if actions appear unsafe, unjustified, or misaligned with mission goals. This dual-mode oversight model ensures that human operators are not only consulted when needed but remain continually empowered to oversee and correct unwanted autonomous behavior.

\paragraph{Autonomy with Accountability} Together, these cognitive guardrails enable CAIRN to support principled autonomy: sUAS agents can reason, adapt, and act within mission constraints while remaining accountable to human oversight and bounded by safety constraints.
\section{Validation}
\label{sec:implementation}
We developed a custom simulation environment that supports real-time interaction, belief updates, and adaptive decision-making, and used it for initial development and validation of the CAIRN framework. In addition, we are current integrating CAIRN into our existing DroneResponse physical-world platform. 

\subsection{Simulation Environment}
The CAIRN simulation was implemented using Python,  PyQt5 for the interface, GPT-4o for natural language reasoning, and pgmpy for discrete Bayesian model construction and probabilistic inference. The structure of our Bayesian Model is shown in Figure \ref{fig:bayes_model}, with conditional probabilities (i.e., CPTs) established according to documented behaviors of lost persons, with concrete values mapping to Equations \ref{eq:1}, \ref{eq:2}, and \ref{eq:3} based on our best judgment. Over time, we intend to refine these based on an ongoing series of focal groups with emergency responders (e.g., \cite{next-gen-drones}) and results from field-based experimentation.  The simulation we developed for CAIRN included diverse sUAS maneuvers (e.g., fly to waypoints, circle objects, and follow trails or lakeshores), but excluded normal behaviors such as takeoff, trajectory smoothing, and collision avoidance, all features present in our higher-fidelity simulation environment. These simplifications yielded speed-ups of 50× or more, allowing fast iteration on model development, decision-making logic, and cognitive guardrail behavior.

Experiments were structured around a library of seven lost-person profiles, 20 clues, and three different geographical locations.  For each geographical location we retrieved the satellite map and USGS terrain models, and used a combination of Computer Vision and model analysis to dynamically annotate map features. Features were color-coded for forested areas, shrub-land, waterways, paths, buildings, and shorelines etc. Each test scenario was comprised of a geographical region, weather and lighting conditions, a lost-person profile including their last-known-point (LKP), two to five relevant clues, and two to five non-relevant ones.  To construct the test scenario, we strategically placed relevant clues into the scenario according to viable paths that might reasonably be taken by the lost person, while non-relevant clues were scattered within and around the area covered by the relevant clues. To accelerate testing, we preprocessed each clue's image in advance using GPT-4o and tagged each clue with the resulting natural language description. 

\begin{figure}[t]
\centering
\includegraphics[width=\linewidth]{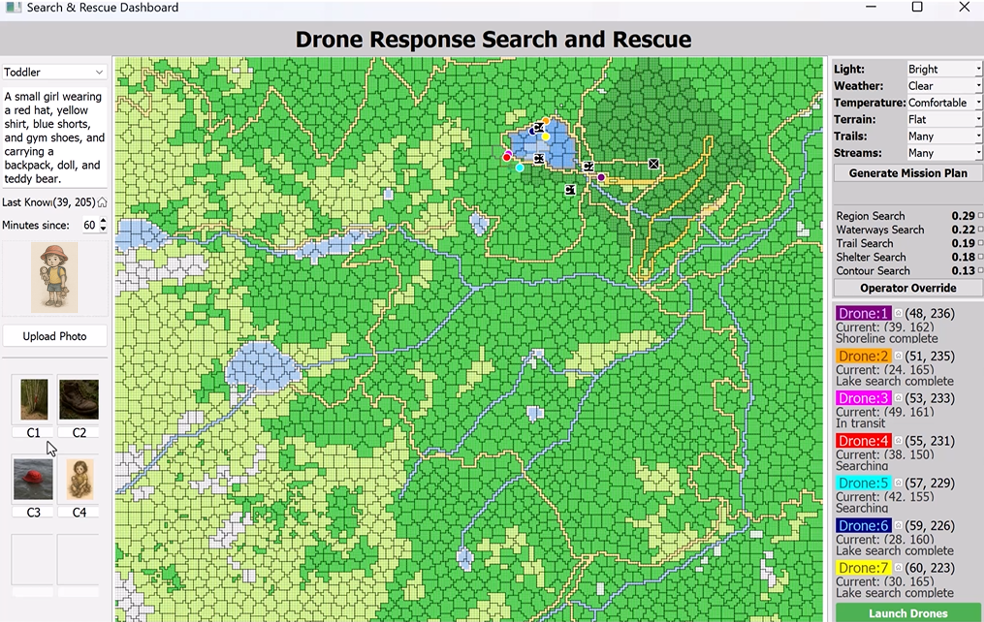}
\caption{A swarm-based search is conducted for a lost child in the Rocky Mountains. The scene is dynamically reconstructed from satellite imagery and terrain models and seven sUAS are dispatched on the search. Through the course of the mission, several clues are identified leading to the final water search, in which the child was found on the lakeshore.  See \url{https://tinyurl.com/DroneSwarm-CAIRN} for a short video }
\label{fig:rockies}
\end{figure}

Figure~\ref{fig:rockies} illustrates one of our test scenarios in which a search for a lost girl takes place near a lake in the Rocky Mountains.  Sixty minutes have elapsed since her last known position at a trailhead parking lot, and seven sUAS have been dispatched on the search. At the start of the mission, the girl's description, along with current weather and terrain conditions, are fed into the Bayesian model, which then identifies {\it Region Search} as the most likely search strategy. The sUAS explore the area near the LKP and soon detect two clues: a doll and a pair of spectacles. The LLM pipeline deems the spectacles irrelevant but highlights the doll as strongly relevant (see Table~\ref{tab:llm_trace_redhat}). This reinforces belief in the current strategy, and {\it Region Search} remains dominant.

\begin{figure}[t]
\centering
\includegraphics[width=.65\linewidth]{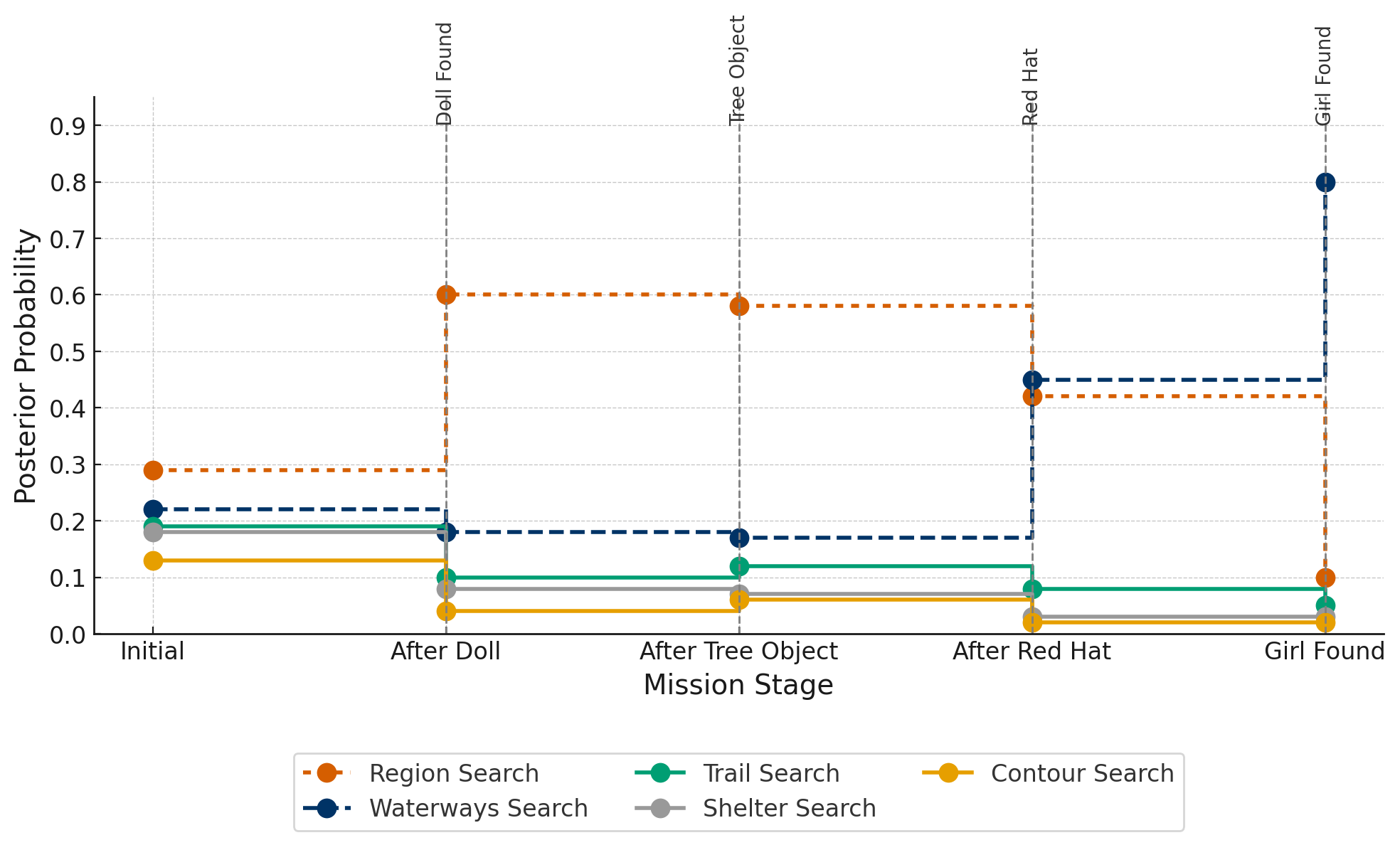}
\caption{Discrete posterior belief updates in response to four major events. The shift from Region to Waterways Search is driven by a high-confidence red hat clue followed by visual confirmation of the child.}
\label{fig:posterior_probs}
\end{figure}

Later, one sUAS spots a red object in a nearby tree. The estimated inspection cost is only 10 minutes, and the object color aligns with the girl's known attire. The Local Mission Planner dispatches a drone to inspect the object, which is ultimately rejected as non-relevant. Shortly afterward, another clue is detected along the lake shoreline and identified as a red hat with high relevance. This new evidence substantially increases belief in {\it Waterways Search}, though not enough to automatically override the existing strategy. As a result, a mission alert is raised, and the strategy transition is confirmed by a human operator. One sUAS is immediately dispatched to search the lakeshore, while three begin a sweep of the lake. Soon after, one sUAS visually confirms the presence of a child, playing near the lake, and matching the lost girl's description. Human responders are notified and the child is successfully rescued. The increase and decrease of beliefs in each search strategy are shown for this mission in Figure \ref{fig:posterior_probs}.

We ran the simulation on a Windows 11 Pro workstation (Build 26100) with an Intel 64-bit processor (2.1 GHz), 64 GB RAM, and a Micro-Star International PRO H610 DP180 motherboard, and used a centralized CAIRN pipeline (mimicking the case that sUAS issue requests over MeshRadio and MQTT to a centralized service on the GCS or in the cloud). In this setup, each stage of the pipeline using GPT-4o took from 30-45 seconds to run, and the GUI displayed results incrementally as they were processed.  We observed expected behaviors including initialization of the Bayesian Model's strategies according to the environment and person profile, correct processing of clues by the LLM, and appropriate adaptations. For example, in the case of the small girl, clues such as a pair of old adult boots and a bicycle were correctly deemed non-relevant, red cloth in a tree triggered an inspection, and a red hat in the water triggered both a systematic search of the lake and a search of the lake-shore, which resulted in finding the lost child.  A replay of this mission is available online at \url{https://tinyurl.com/DroneSwarm-CAIRN}.

\subsection{Towards Physical-World Deployment}
\label{sec:integration}
Our DroneResponse system, provides a robust and proven platform for autonomous multi-sUAS swarming \cite{drone_response_quarry_demo}, and is well suited for deploying the CAIRN framework. It includes a Ground Control Station (GCS) composed from an extensible set of microservices that currently includes an air-leaser, runtime monitoring, and smart mission planning (SMP) \cite{alarcon2024land}. System-wide communication between sUAS, ground-based services, and humans is supported by a MeshRadio which enables both centralized coordination and the ad-hoc formation of coordinated sUAS swarms.  Each sUAS is supported by onboard services, currently including a mission-level state machine, a CV pipeline, and runtime anomaly detection.  The entire ecosystem is deployable using simulated and/or physical sUAS, and supports human interaction via diverse Hardware and Graphical User Interfaces.

Our novel Edge Matrix Cube (mCube), depicted in Figure \ref{fig:drones}, provides a plug-and-play hardware and software stack, compatible with both Ardupilot and PX4 flight controllers. It hosts our own DroneResponse autonomous pilot software on a Jetson Orin NX, delivering up to 100 TOPS (Tera Operations Per Second) of AI performance and providing ample processing power for real-time onboard AI, computer vision, and diverse reasoning tasks. It also incorporates mesh radio connectivity, diverse sensors (e.g., cameras), and infrastructure for cooling and power management. The mCubes have been flight-tested on the Aurelia Max 6  and the Inspired Flight IF 1200A, which are capable of 60 and 40 minutes of flight respectively at distances of up to 4 miles.

The DroneResponse architecture readily encompasses the CAIRN infrastructure. As we move forward with our integration plans, CAIRN’s Local Planner will be deployed on the mCube to enable context-aware reasoning at the edge with access to system-wide services. The Global Planner, Bayesian model, and LLM-based reasoning pipeline will be hosted on the Ground Control Station (GCS), and human interaction components added as extensions to existing GUIs. We are aiming for full deployment of CAIRN for physical world deployment by the Fall of 2025, thereby enabling open-world reasoning and adaptive mission planning for search and rescue in uncertain environments.
\section{Conclusion}
\label{sec:conclusion}

The CAIRN framework represents a novel approach for enabling principled, autonomous adaptation in open-world environments by combining Bayesian belief updates, contextual reasoning via an LLM pipeline, and cognitive guardrails that ensure safe, transparent, and interpretable decision-making. Unlike prior approaches that rely on predefined event sets~\cite{chenini2019embedded, fan2024uav, Lin2010}, CAIRN supports dynamic, mission-level decisions under uncertainty.

CAIRN was designed with input from white papers; however, we are engaging in an ongoing participatory design process with SAR experts to validate the SAR knowledge-base, determine whether decisions made by the sUAS align with their expectations, and to assess whether the human-input triggers match their situational awareness needs.  For our initial proof-of-concept purposes, many of the system’s parameters, such as Bayesian probabilities and evidence update factors, were defined manually by our own team members, while prompts used in CAIRN's LLM were established through systematic testing. Further experimentation with diverse prompts, and different weightings for relevance and confidence, is therefore needed to further refine the underlying models.
Broader evaluation across randomized missions, comparisons with traditional methods such as grid-based search, and findings from physical world deployments, will also be explored in future work. 

Finally, while we developed CAIRN with a focus on search and rescue (SAR), it is also highly applicable to other domains that require real-time reasoning and adaptation. We are developing infrastructure for deploying it in projects such as solar farm inspections, pond-weed surveillance, wildfire management, and deforestation mapping. In these contexts, sUAS must detect anomalies, patterns, or objects of interest, interpret observations, and adapt behavior accordingly.

\begin{acks}
Preliminary ideas in this paper were presented in a keynote address at the 20th International Conference on Software Engineering for Adaptive and Self-Managing Systems in May 2025. Funding for foundational aspects of this project was provided by the USA National Science Foundation under grant \# 1931962.
\end{acks}

\bibliographystyle{ACM-Reference-Format}
\bibliography{decisions}




\end{document}